\documentclass[runningheads]{llncs}
\usepackage{graphicx}
\usepackage{amsmath,amssymb} 
\usepackage{color}
\usepackage{bm}
\usepackage{tabularx}
\usepackage{multirow}
\usepackage{hhline}
\usepackage[caption=false]{subfig} 
\usepackage{booktabs}
\usepackage{sidecap}
\usepackage{rotating}
\usepackage{lipsum}
\usepackage[normalem]{ulem}
\usepackage{float}
\usepackage{wrapfig}
\usepackage{xcolor}
\usepackage{xspace}
\usepackage{authblk}
\usepackage{adjustbox}
\usepackage{moresize}
\usepackage[symbol]{footmisc}
\usepackage[pagebackref=true,breaklinks=true,letterpaper=true,colorlinks,bookmarks=false]{hyperref}

\def\etal{\emph{et al.}\xspace}
\def\ie{\emph{i.e.}\xspace}
\def\eg{\emph{e.g.}\xspace}

\newcommand{\absent}{\texttt{absent}\xspace}
\newcommand{\present}{\texttt{present}\xspace}

\begin{document}
\title{Long-term Tracking in the Wild: A Benchmark}

%


\author{Jack Valmadre\footnote[1]{Equal contribution}$^{,1}$,\quad Luca Bertinetto$^{*,}$\footnote[2]{Work done while at University of Oxford}$^{,1,3}$,\quad Jo\~{a}o F.\ Henriques$^1$,\quad Ran Tao$^2$,\vspace{0.12cm} \newline Andrea Vedaldi$^1$,\quad Arnold Smeulders$^2$,\quad Philip Torr$^{1, 3}$,\quad Efstratios Gavves$^{*,2}$}
\institute{$^1$ University of Oxford\quad$^2$ University of Amsterdam \quad$^3$ FiveAI} 

%
\authorrunning{Valmadre, Bertinetto, Henriques, Tao, Vedaldi, Smeulders, Torr, Gavves}
%

%
\maketitle              
%
\begin{abstract}
\noindent We introduce the OxUvA dataset and benchmark for evaluating single-object tracking algorithms.
Benchmarks have enabled great strides in the field of object tracking by defining standardized evaluations on large sets of diverse videos.
However, these works have focused exclusively on sequences that are just tens of seconds in length and in which the target is always visible.
Consequently, most researchers have designed methods tailored to this ``short-term'' scenario, which is poorly representative of practitioners' needs.
Aiming to address this disparity, we compile a long-term, large-scale tracking dataset of sequences with average length greater than two minutes and with frequent target object disappearance.
The OxUvA dataset is much larger than the object tracking datasets of recent years: it comprises 366 sequences spanning 14 hours of video.
We assess the performance of several algorithms, considering both the ability to locate the target and to determine whether it is present or absent.
Our goal is to offer the community a large and diverse benchmark to enable the design and evaluation of tracking methods ready to be used ``in the wild''.
The project website is \href{http://oxuva.net/}{\nolinkurl{oxuva.net}}.

\end{abstract}


\section{Introduction}\label{sec:intro}
\begin{figure*}[th!]
\centering
\adjustbox{max width=0.9\linewidth}{
\includegraphics[scale=1]{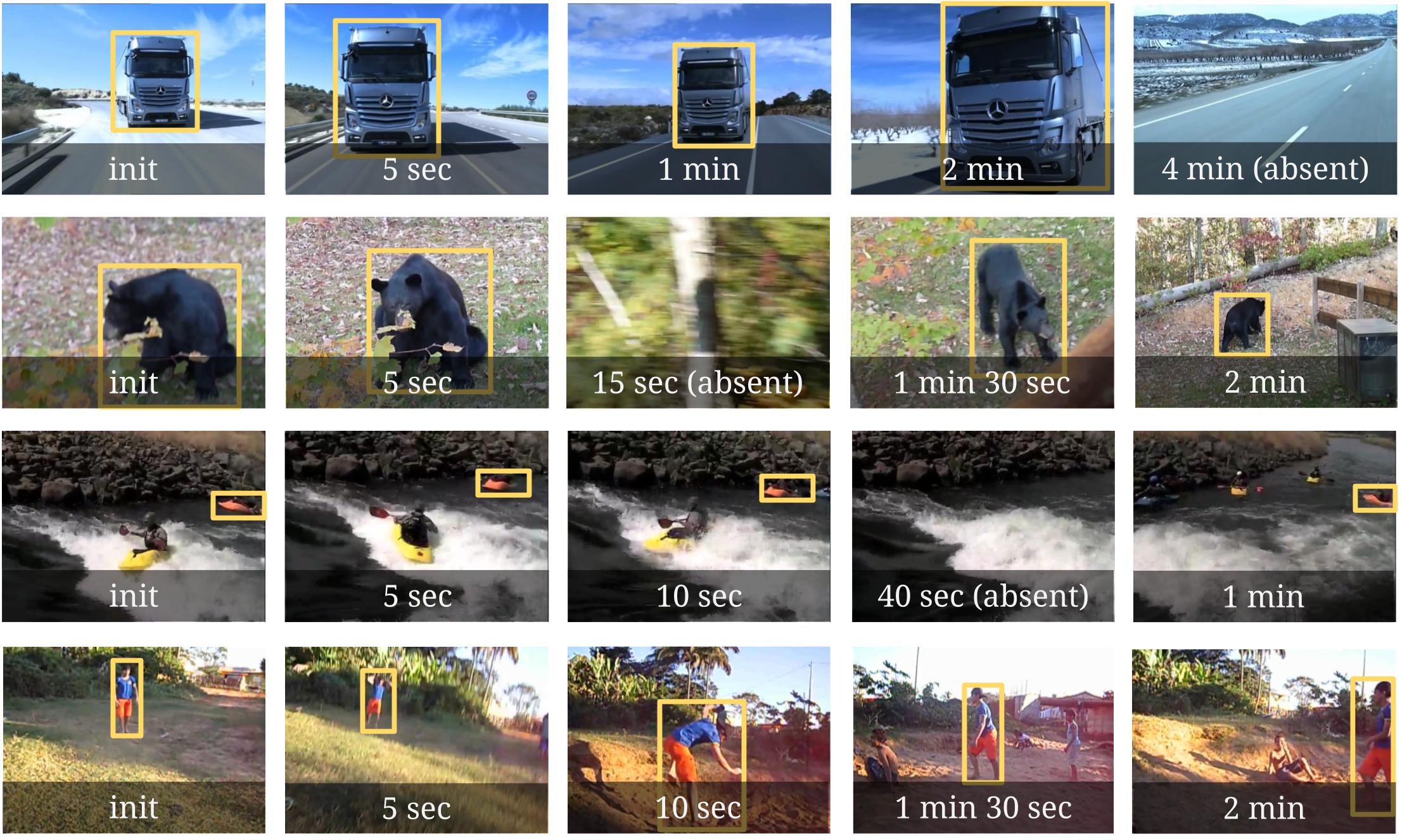}
}
\caption{Example sequences and annotations.
Unlike standard benchmarks, our dataset focuses on long sequences with annotated disappearance of the target object.}
\label{fig:qualitative}
\end{figure*}

Visual object tracking is the task of locating an arbitrary, user-specified target in all frames of a video sequence.
Traditionally, the target is specified using a rectangle in a single frame.
The ability to track an arbitrary object would be useful for many applications including video analytics, surveillance, robotics, augmented reality and video editing.
However, the requirement to be able to track \emph{anything} given only a single example presents a significant challenge due to the many complex factors that affect appearance, such as out-of-plane rotation, non-rigid deformation, camera perspective, motion blur, illumination changes, occlusions and clutter.

Tracking benchmarks~\cite{smeulders2014visual,wu2015object,kristan2016novel,liang2015encoding,li2016nus,mueller2016benchmark} have played a huge role in the advancement of the field, enabling the objective comparison of different techniques and driving impressive progress in recent years.
However, these benchmarks have focused on the problem of ``short-term tracking'' according to the definition of Kristan \etal~\cite{kristan2016novel}, which does not require methods to perform re-detection.
This implies that the object is always present in the video frame.

This constraint was perhaps introduced with the intention of limiting the scope of the problem to facilitate progress.
However, the influence of these benchmarks has been so pervasive that the large majority of modern trackers estimate a bounding box in \emph{every frame}, implicitly assuming that the target never disappears from the scene.
For most practical applications, however, it is critical to track objects through disappearance and re-appearance events, and further, to be \emph{aware} of the presence or absence of the object.

Existing benchmarks are also \emph{short-term} in the literal sense that the average video length does not exceed 20-30 seconds.
Such short sequences do not accurately represent practical applications, in which videos can easily be several minutes, and possibly arbitrarily long.
Little is known of which trackers are most effective in this scenario: while short-term benchmarks make a particular effort to include a variety of challenging situations, tracking in long videos may introduce unforeseen challenges.
For instance, many methods use their past predictions to update an internal appearance model.
While this generally improves the results in short-term tracking, the accumulation of errors over time leads to model drift~\cite{santner2010prost}, which may have a catastrophic effect in longer videos.

With this work, we introduce a novel single-object tracking benchmark and aim to advance the literature through several contributions:
\begin{enumerate}
\item Our dataset contains sequences with an average duration of 2.4 minutes, seven times more than OTB-100.
With 14 hours of video (1.5 million frames), it is also the largest tracking dataset to date.
\item We deliberately assess methods in situations where the target disappears, an event that occurs in roughly half the videos of the dataset.
\item Unlike existing tracking benchmarks, we split the data into two sets: development (\emph{dev}) and \emph{test}.
The ground-truth for the test set is only accessible via a rate-limited evaluation server.
This helps avoid over-fitting hyper-parameters to the singular dataset of the benchmark, thus promoting generalization.
\item We design a new evaluation that captures the ability of a tracker to both decide the presence or absence of the object and to locate it in the image.
\item Instead of manually-annotated binary attributes, which can be subjective, we propose \emph{continuous attributes}, which allow an in-depth study of how smoothly-varying conditions affect each tracker.
\item We evaluate and compare several representative methods from the literature that either perform well or seem particularly well-suited to the problem.
\end{enumerate}
We hope this paper encourages the community to relax the strong assumptions of short-term tracking benchmarks and to develop methods that can be readily used in the many applications that present a ``long-term'' scenario.

\section{Related Work}\label{sec:related}
\paragraph{Large-scale video datasets.}
There has been an increasing interest by the computer vision community in large-scale video datasets.
Two notable examples are the datasets for object detection in video, ImageNet VID~\cite{ILSVRC15} and YouTube Bounding Boxes~\cite{real2017youtube} (YTBB).
ImageNet VID contains 20 classes and almost four thousand videos, with every object instance annotated in every frame.
YTBB contains 23 classes and 240k videos from YouTube, with a single instance of each class annotated once per second for up to twenty seconds.
YTBB specifically aims to comprise videos ``in the wild'' by considering only those with 100 views or less on YouTube.
This was observed to be a good heuristic for selecting unedited videos of personal users.
This work uses YTBB as a source from which to curate and further annotate the sequences that constitute our long-term benchmark.

\paragraph{Tracking benchmarks.}
The practice of evaluating tracking algorithms has improved considerably in recent years.
In the past, researchers were limited to evaluating tracking performance on a mere handful of sequences (\eg \cite{ross2008incremental,babenko2009visual,mei2009robust,breitenstein2009robust}).
Benchmarks like ALOV~\cite{smeulders2014visual}, VOT~\cite{kristan2016novel} and OTB~\cite{wu2015object} underlined the importance of testing methods on a much larger set of sequences which encompasses a variety of object classes and factors of variation.
To evaluate tracker performance, ALOV computes an F-score per video using a 50\% intersection-over-union (IOU) criterion, then visualizes the distribution of F-scores.
OTB instead reports, for a range of thresholds, the percentage of frames in which the IOU exceeds each threshold.
The VOT benchmark is distinct from others in that trackers are restarted after each failure.
Motivated by a correlation study, two metrics (mean IOU and number of failures) are used to quantify tracker performance, and these are jointly expressed in the Expected Average Overlap.
Recently, TempleColor~\cite{liang2015encoding} (TC), UAV123~\cite{mueller2016benchmark} and NUS-PRO~\cite{li2016nus} have introduced new sequences and adopted the OTB performance measures.

Differently from our work, standard benchmarks only offer sequences that are relatively short (lasting 7-30 seconds on average) and do not contain disappearance of the target, thus not requiring methods to perform re-detection.
In the rare frames where the object is fully occluded, OTB-100 places a bounding box on top of the occluder, while UAV123 ignores the frame during evaluation.

\paragraph{Long-term tracking.}
To our knowledge, the first attempt in the literature to evaluate tracking algorithms on long sequences with disappearances was the \textit{long-term detection and tracking} workshop (LTDT)~\cite{LTDT}.
Despite the fact that the number of frames in LTDT is comparable to OTB-100~\cite{wu2015object}, its modest number of sequences (five) makes it unsuitable for assessing the performance of a general purpose tracker.
Tao \etal~\cite{tao2017tracking} investigated object tracking in half-hour sequences using the periodic, symmetric extension of short sequences.
However, this does not necessarily capture the same level of difficulty as real videos.

Two long-term tracking datasets have been proposed in concurrent work~\cite{moudgil2017long,lukezic2018now}, both of which include sequences with labelled target absences.
However, to our knowledge, neither provides a test set with secret ground-truth.

\section{Long-Term Tracking Dataset}\label{sec:definition-dataset}
\subsection{Dataset Compilation and Curation}
Our aim is to collect long and realistic video sequences in which the target object can disappear and re-appear.
We use the YTBB~\cite{real2017youtube} \textit{validation} set as a superset from which to select our data.
YTBB contains 380k tracklets from 240k different YouTube videos, annotated at 1Hz with either a bounding box or the absent label.
Despite being an excellent starting point, the data of YTBB are not ready to be used for the purpose of evaluating methods in a long-term scenario.
Several stages of manual data curation are required.

The major issue is that the tracklet duration is limited to less than 20 seconds.
However, it often occurs that multiple tracklets in one video refer to the same object instance.
We identify these tracklets and combine their annotations in order to obtain significantly longer sequences, albeit with large gaps between annotated segments.
This process involves finding the videos which contain multiple tracklets of the same class, watching the video and manually specifying which (if any) refer to the same object instance.
Another issue with YTBB is that the first frame of a track may not be a suitable initial example to specify the target.
To remedy this issue, for each video we manually select the first annotated frame in which the bounding box alone provides a clear and sufficient definition of the target.
All annotations preceding this frame are discarded.
The final manual stage is to exclude sequences that are of little interest for tracking, for example those in which the target object undergoes little motion or fills most of the image in most of the frames.
To ensure the quality of annotations, all manual operations have been performed by a pool of five expert annotators.
Each sequence has been assessed by two annotators and included only if both agreed.

Once this manual process was complete, we assessed the performance of a naive baseline that simply reports the initial location in every subsequent frame.
We then discarded all sequences in which this trivial tracker achieves at least 50\% IOU in at least 50\% of the frames. 

\begin{table}[t]
\centering
\caption{Comparison of the proposed OxUvA long-term tracking benchmark to existing benchmarks.
Our proposal presents the longest average sequence length and is the only one testing trackers against object disappearance.
}
\label{table:comparison}
\adjustbox{max width=\linewidth}{
\begin{tabular}{l @{\hskip 2ex} c @{\hskip 2ex} c @{\hskip 2ex} c @{\hskip 2ex} c @{\hskip 2ex} c @{\hskip 2ex} c @{\hskip 2ex} c @{\hskip 2ex} c @{\hskip 2ex} c}
\toprule
& OxUvA & OTB-100 & VOT & UAV123 & DTB & NUS-PRO & TC & ALOV & NfS \\
& 2018 & 2015 \cite{wu2015object} & 2017 \cite{kristan2016novel} & 2016 \cite{mueller2016benchmark} & 2017 \cite{li2017visual} & 2016 \cite{li2016nus} & 2015 \cite{liang2015encoding} & 2013 \cite{smeulders2014visual} & 2017 \cite{kiani2017need} \\
\midrule
Frames &\textbf{1.55M} & 59k & 21k & 113k & 15k & 135k & 60k & 152k & 380k \\
Tracks & \textbf{366} & 100 & 60 & 123 & 70 & \textbf{365} & 128 & 314 & 100 \\
\dots w/ \absent{} labels & \textbf{52\%} & 0\% & 0\% & 0\% & 0\% & 0\% & 0\% & 0\% & 0\% \\
Avg length (min)& \textbf{2.36} & 0.33 & 0.20 & 0.51 & 0.12 & 0.21 & 0.26 & 0.27 & 0.26 \\
Median length (min) & \textbf{1.46} & 0.22 & 0.17 & 0.49 & 0.10 & 0.17 & 0.22 & 0.15 & 0.17 \\
Max length\ (min) & \textbf{20.80} & 2.15 & 0.83 & 1.71 & 0.35 & 2.8 & 2.15 & 3.32 & 1.44 \\
Min length (min) & \textbf{0.50} & 0.04 & 0.02 & 0.06 & 0.04 & 0.08 & 0.04 & 0.01 & 0.01 \\
Avg \absent{} labels & \textbf{2.2} & 0 & 0 & 0 & 0 & 0 & 0 & 0 & 0 \\
Object classes & 22 & 16 & 24 & 9 & 15 & 8 & \textbf{27} & -- & -- \\
\bottomrule
\end{tabular}
}
\end{table}

Our final dataset comprises 366 object tracks in 337 videos.
These were selected from an initial pool of about 1700 candidate videos, all of which were watched by at least two expert annotators.
Table~\ref{table:comparison} summarizes some interesting statistics and compares the proposed dataset against existing ones.
Remarkably, the total number of frames is respectively 26 and 10 times larger than the popular OTB-100 and ALOV respectively, making our proposed dataset the largest to date.
Moreover, existing benchmarks never label the target object as absent.
In contrast, our proposal contains an average of 2.2 \absent{} labels per track and at least one labelled disappearance in 52\% of the tracks.
Finally, the sequences we propose are much longer, exhibiting an average duration of 2.3 minutes.

\subsection{Data Subsets and Challenges}

We split our dataset of 366 tracks into \textit{dev} and \textit{test} sets of 200 and 166 tracks respectively.
The classes in the dev and test sets are disjoint, and this split is chosen randomly.
The dev set contains bear, elephant, cat, bus, knife, boat, dog and bird; the test set contains zebra, potted plant, airplane, truck, horse, cow, giraffe, person, bicycle, umbrella, motorcycle, skateboard, car and toilet.
The ground-truth labels for the testing set are secret, and can only be accessed through the evaluation server\footnote{\url{https://competitions.codalab.org/competitions/19529}}.
All results in the main paper are for the test set unless otherwise stated.
A comparison between the dev and test sets can be found in the supplementary material.

Using these subsets, we further define two challenges: \textit{constrained} and \textit{open}.
For the constrained challenge, trackers can be developed using only data from our dev set (long-term videos), from the dev classes in the original YTBB \textit{train} set and from standard tracking benchmarks (see the website for precise rules).
For the open challenge, trackers can use any public dataset except for the YTBB \textit{validation} set, from which OxUvA is constructed.
The constrained setting is closer to traditional model-free or one-shot tracking, since the object categories in the test set have not been seen before.
All trackers in the constrained challenge are automatically entered into the open challenge.
Note that methods using model parameters that were pre-trained for an auxiliary task are only eligible for the open challenge.
The results for the constrained trackers alone are deferred to the supplementary material.

\subsection{Annotation Density}
\label{sec:task}

\begin{figure}[t]
\centering
\adjustbox{max width=\linewidth}{
\subfloat[Fixed number of videos (100)]{
    \includegraphics[width=55mm]{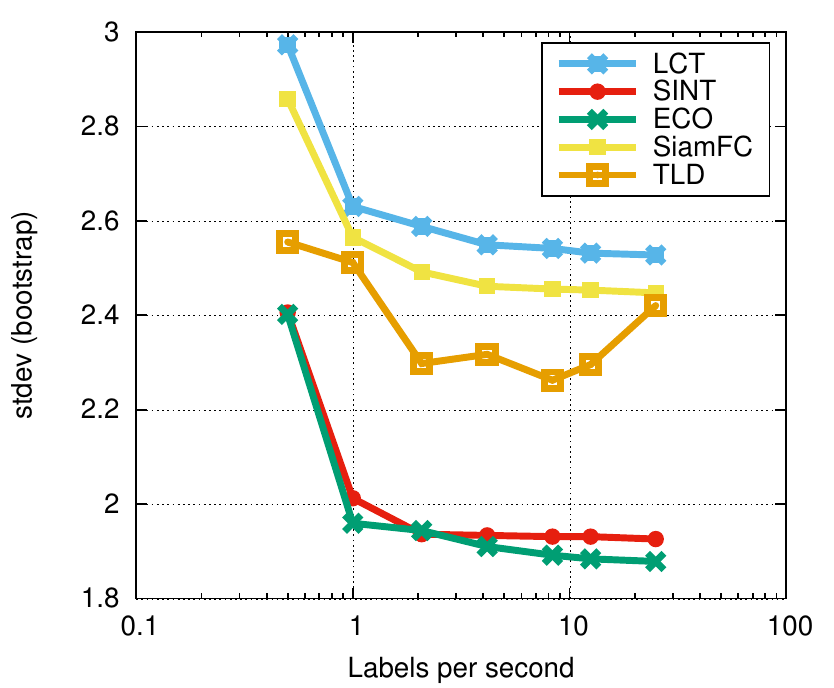}}
\subfloat[Fixed label frequency (1Hz)]{
    \includegraphics[width=55mm]{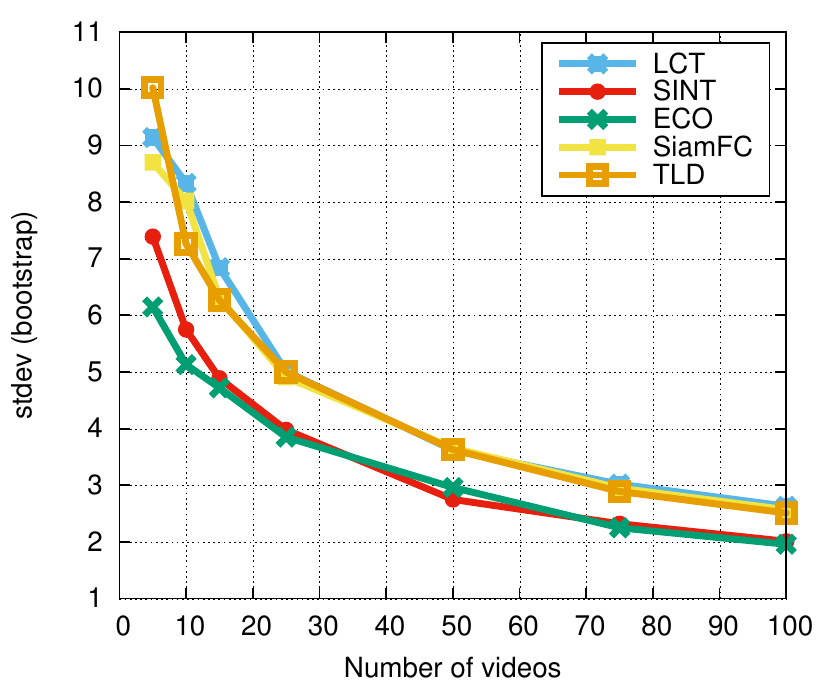}}
\subfloat[Fixed label budget]{
    \includegraphics[width=55mm]{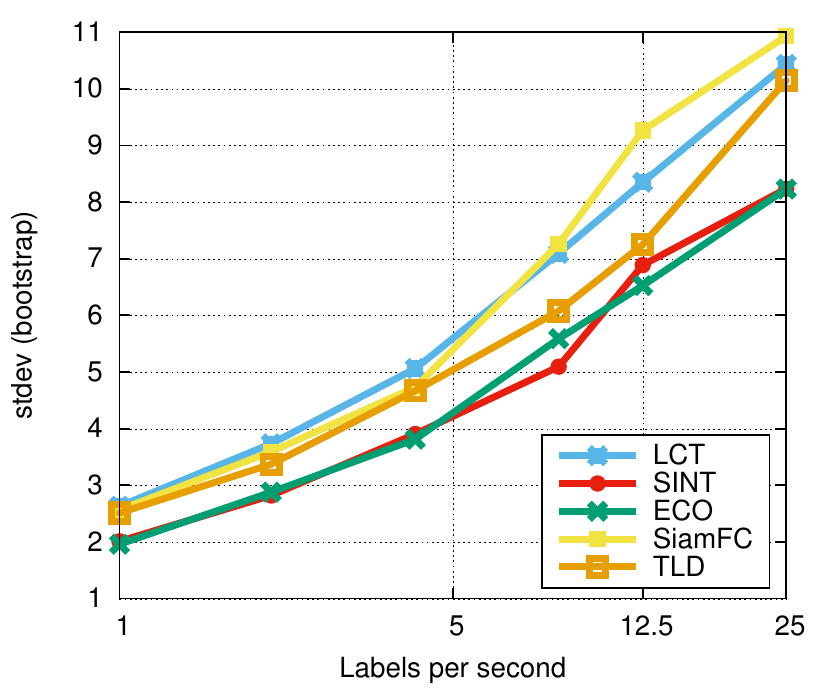}}
}
\caption{Impact of annotation density and number of sequences on the evaluation reliability (higher standard deviation implies a less reliable evaluation).}
\label{fig:annotation-density}
\end{figure}

Unlike most existing tracking benchmarks, in which every frame is labelled, the tracklets in YTBB are only labelled at a frequency of 1Hz.
We argue that this is sufficient for tracker evaluation since \emph{a)} it is unlikely that a tracker will fail and recover within one second, and \emph{b)} a tracking failure of less than a second would be relatively harmless in many applications.
To verify this hypothesis, we investigate the results of several representative trackers on the OTB-100~\cite{wu2015object} benchmark, varying the label frequency and number of videos in three experiments.

We study the effect of each experiment on the variance of the overall score considering the test set to be a random variable.
Lower variance indicates a more reliable evaluation.
Although we only have one sample from the distribution of test sets, this distribution can be approximated by repeatedly bootstrap sampling the one available test set~\cite{wasserman2013all}.
We adopt the AUC score as our performance measure and use the One Pass Evaluation protocol of OTB-100.

{\bf Experiment 1:} \emph{Vary the label frequency from 0.5 to 25Hz, keeping the number of videos fixed at 100.} (Fig.~\ref{fig:annotation-density}a)
With a fixed number of videos, a higher labelling density only marginally improves reliability.
In fact, between 1Hz and 25Hz, we did not observe a significant difference in standard deviation.
A meaningful degradation only occurs at 0.5Hz.

{\bf Experiment 2:} \emph{Vary the number of videos from 5 to 100, keeping the label frequency fixed at 1Hz.} (Fig.~\ref{fig:annotation-density}b)
Increasing the number of videos while keeping the frequency constant results in a steady and significant reduction in variance.

{\bf Experiment 3:} \emph{With a fixed budget of labels for the dataset, increase the label frequency by decreasing the number of videos (from 100 videos at 1Hz to 4 videos at 25Hz).} (Fig.~\ref{fig:annotation-density}c)
A more reliable evaluation is obtained by increasing the number of videos at the expense of having fewer labels per second.
Annotating \emph{more} videos \emph{sparsely} (at 1Hz) leads to $4$-$5\times$ smaller standard deviation than annotating \emph{fewer} videos \emph{densely} (at 25Hz).

We conclude that \emph{a)} labelling at 1Hz does not adversely affect the robustness of evaluation and \emph{b)} a large number of videos is paramount.

\section{Tracker Evaluation}\label{sec:evaluation}
\subsection{Evaluating Object Presence and Localization}

Given an initial bounding box for the target, we require a tracker to predict either \present{} or \absent{} in each subsequent frame, and to estimate its location with an axis-aligned bounding box if present.
This raises the question of how to evaluate a tracker's ability both to locate the target and to decide its presence.

With this intention, we introduce an analogy to binary classification.
Let us equate object presence with the positive class and absence with the negative.
In a frame where the object is absent, we declare a true negative (TN) if the tracker predicts \absent{}, and a false positive (FP) otherwise.
In a frame where the object is present, we declare a true positive (TP) if the tracker predicts \present{} \emph{and} reports the correct location, and a false negative (FN) otherwise.
The location is determined to be correct if the IOU is above a threshold.
Using these definitions, we can quantify tracking success using standard performance measures from classification.

However, some performance measures are inappropriate because the dataset possesses a severe class imbalance: although target disappearance is a frequent event, and occurs in roughly half of all sequences, only 4\% of the actual annotations are \absent{}.
As a result, it would be possible to achieve high accuracy, high precision and high recall without making a single \absent{} prediction.
We therefore propose to evaluate trackers in terms of True Positive Rate (TPR) and True Negative Rate (TNR), which are \emph{invariant} to class imbalance~\cite{fawcett2006introduction}.
TPR gives the fraction of present objects that are reported \present{} and correctly located, while TNR gives the fraction of absent objects that are reported \absent{}.
Note that, in contrast to typical binary classification problems, these metrics are not symmetric.
While it is trivial to achieve $\text{TNR} = 1$ by reporting \absent{} in every frame, it is only possible to achieve $\text{TPR} = 1$ by reporting \present{} in every frame \emph{and} successfully locating the object.

To obtain a single measure of tracking performance, we propose the geometric mean $\text{GM} = \sqrt{\text{TPR} \cdot \text{TNR}}$.
This has the advantage that relative improvements in either metric are equally valuable since $\sqrt{(\alpha x) y} = \sqrt{x (\alpha y)}$.

\subsection{Operating Points}\label{sec:evaluation:operating-points}

In the object detection literature, it is usual to report a precision-recall curve, which plots the range of operating points that are obtained by varying a threshold on the scores of the predictions (\ie to decide which are considered detections).
The overall performance is then computed from multiple operating points, typically the average precision at multiple desired values of recall.
Unfortunately, we cannot use the same methodology because trackers are \emph{causal}.
If we were to evaluate trackers using a range of operating points that are obtained without re-running the tracker, it may give an artificial advantage to state-less algorithms.
Furthermore, if the tracker maintains an internal state, applying a threshold would cause its reported state to diverge from its internal state.
Therefore, we require the tracker to output a hard decision in each frame, corresponding to a single point in TPR-TNR space.

However, even without making use of prediction scores, we can still consider a simple range of operating points.
Specifically, a TPR-TNR curve is obtained by randomly flipping each \present{} prediction to \absent{} with probability $p \in [0, 1]$.
This traces a straight line to the trivial operating point $\text{TPR} = 0$, $\text{TNR} = 1$, at which all predictions are \absent{} (see Figure~\ref{fig:main}, left).
This line establishes a lower bound on the TPR of a method at a higher TNR.
One tracker is said to be dominated by another if its TPR is below the lower bound of the other tracker at the same TNR.

Since most existing trackers never predict \absent{}, they will have $\text{GM} = \text{TNR} = 0$.
To enable a more informative comparison to these trackers, we instead consider the maximum geometric mean along this lower bound
\begin{equation}
\text{MaxGM} = \max_{0 \le p \le 1} \sqrt{((1-p) \cdot \text{TPR}) ((1-p) \cdot \text{TNR} + p)} \enspace .
\end{equation}

\section{Evaluated Trackers}\label{sec:trackers}
We now explore how methods from the recent literature perform on our dataset.
We limit the analysis to a selection of ten baselines which have shown strong performance or have affinity to the scenario we are considering.
The baselines we select are roughly representative of three groups of methods.
\begin{itemize}
\item We first consider \textbf{LCT}~\cite{LCT},  \textbf{EBT}~\cite{EBT} and \textbf{TLD}~\cite{TLD}, three methods that have an affinity with long-term tracking for their design.
Although based on different features and classifiers, they are each capable of locating the target anywhere in the frame, an important property when the target can disappear.
This is in contrast to most methods, which search only a local neighbourhood.
Unfortunately, EBT does not output the presence or absence of the object, and its source code is not available.
\item As a second family, we consider methods that originate from short-term correlation filter trackers like KCF~\cite{henriques2015high}.
In particular, we chose recent methods which can operate in real-time and achieve high performance: \textbf{ECO-HC}~\cite{ECO}, \textbf{BACF}~\cite{kiani2017learning} and \textbf{Staple}~\cite{bertinetto2016staple}.
\item Lastly, we consider three popular algorithms based on deep convolutional networks: \textbf{MDNet}~\cite{nam2016learning} and the Siamese network-based trackers \textbf{SINT}~\cite{SINT} and \textbf{SiamFC}~\cite{bertinetto2016fully}.
Both SINT and SiamFC only evaluate the offline-learned similarity function during tracking, whereas MDNet performs online fine-tuning.
SiamFC is fully-convolutional, adopts a five-layer network and it is trained from scratch as a similarity function.
SINT uses RoI pooling~\cite{girshick2015fast}, is based on a VGG-16~\cite{simonyan2014very} architecture pre-trained on ImageNet and fine-tuned on ALOV and uses bounding-box regression during tracking.
\end{itemize}

From the recent literature, TLD and LCT were the only methods that we could find with source code available that determine the presence or absence of the object.
In order to have an additional method with $\text{TNR} \ne 0$, we equipped SiamFC with a simple re-detection logic similar to that described in~\cite{supancic2017tracking}.
If the maximum score of the response falls below a threshold, the tracker enters \emph{object absent} mode.
From this state, it considers a search area at a random location in each frame until the maximum score again surpasses the threshold, at which point the tracker returns to \emph{object present} mode.
Note that this implementation is method agnostic, does not require extra time for re-detection, and can be applied to any method which uses local search and produces a score in every frame.
For both SiamFC and SiamFC+R we used the baseline model from the CFNet paper~\cite{valmadre2017end}.

For all methods, we use the code and default hyper-parameters provided by the authors.
None of the trackers have been trained on YTBB or tuned for our long-term dataset.
However, some models have been trained on external datasets that share classes with YTBB: SINT and MDNet are initialized with networks pre-trained for image classification and SiamFC is trained on ImageNet VID.

\section{Analysis}\label{sec:experiments}
\addtolength{\tabcolsep}{3pt}

\newcommand{\maintable}{
\scriptsize
\begin{tabular}{l c c c c}
\toprule
& fps & TNR & TPR & MaxGM \\
\midrule
SiamFC+R & 52 & 0.481 & 0.427 & 0.454 \\
TLD~\cite{TLD} & 25 & 0.895 & 0.208 & 0.431 \\
LCT~\cite{LCT} & 27 & 0.537 & 0.292 & 0.396 \\
MDNet~\cite{nam2016learning} & 0.5 & 0 & 0.472 & 0.343 \\
SINT~\cite{SINT} & 2 & 0 & 0.426 & 0.326 \\
ECO-HC~\cite{ECO} & 60 & 0 & 0.395 & 0.314 \\
SiamFC~\cite{valmadre2017end} & 52 & 0 & 0.391 & 0.313 \\
EBT~\cite{EBT} & 5 & 0 & 0.321 & 0.283 \\
BACF~\cite{kiani2017learning} & 40 & 0 & 0.316 & 0.281 \\
Staple~\cite{bertinetto2016staple} & 80 & 0 & 0.273 & 0.261 \\
\bottomrule
\end{tabular}
}

\begin{figure}[t]
\centering
\subfloat{\raisebox{-.47\height}{\includegraphics[height=48mm]{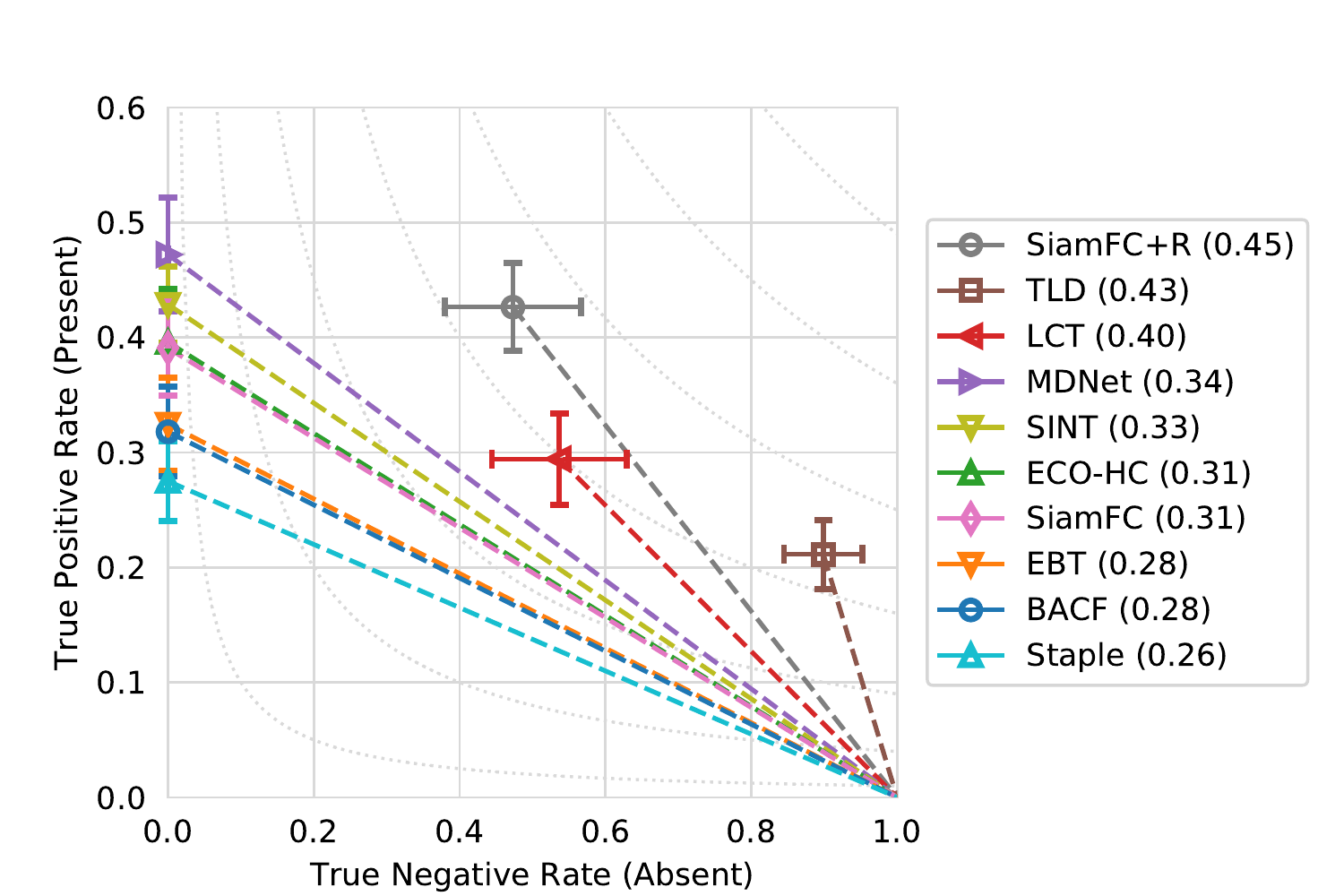}}}
\subfloat{\adjustbox{max width=50mm}{\maintable}}
\caption{Accuracy of the evaluated trackers in terms of True Positive Rate (TPR) and True Negative Rate (TNR) for $\text{IOU} \ge 0.5$.
The figure shows each tracker on a 2D plot (top right is best).
Trackers that always report the object \present{} appear on the vertical axis.
The dashed lines are obtained by randomly switching predictions from \present{} to \absent{}.
Methods are ranked by the maximum geometric mean along this line.
The level sets of the geometric mean are shown in the background.
}
\label{fig:main}
\end{figure}

\paragraph{Main evaluation.}

Figure~\ref{fig:main} (left) shows the operating points of the evaluated methods in a TPR vs.\ TNR plot assuming overlap criterion $\text{IOU} \ge 0.5$.
The exact numbers are detailed in the accompanying table.
Most methods are not designed to report absent predictions, therefore their operating points lie on the vertical axis ($\text{TNR} = 0$).
The dashed lines represent operating points that can be obtained by randomly flipping predictions from \present{} to \absent{} as described in Section~\ref{sec:evaluation:operating-points}.
MDNet, SiamFC+R and TLD dominate other methods in the sense that their collective lower bounds exceed all other trackers.
The following sections will investigate the results in greater depth.

To obtain error-bars, the set of videos is considered a random variable and the variance of each scalar quantity is estimated using bootstrap sampling~\cite{wasserman2013all} as in the earlier experiments.
Naively assuming each variable to be approximately Gaussian, error-bars are plotted for the 90\% confidence interval ($\pm 1.64 \sigma$).
This technique will be used in all following experiments.

\paragraph{Tracker performance over time.}
\label{sec:video_length}

\begin{figure}[t]
\centering
\subfloat{\includegraphics[height=48mm]{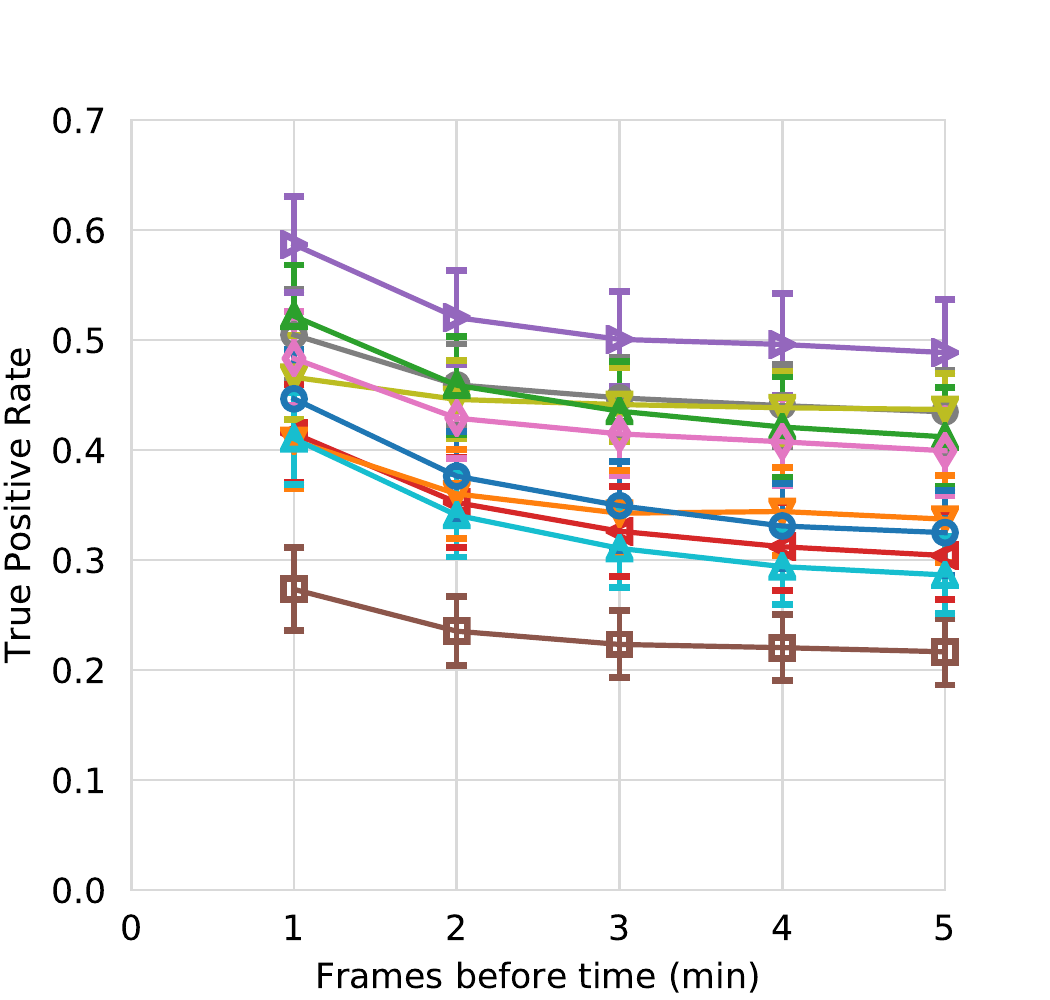}}
\subfloat{\includegraphics[height=48mm]{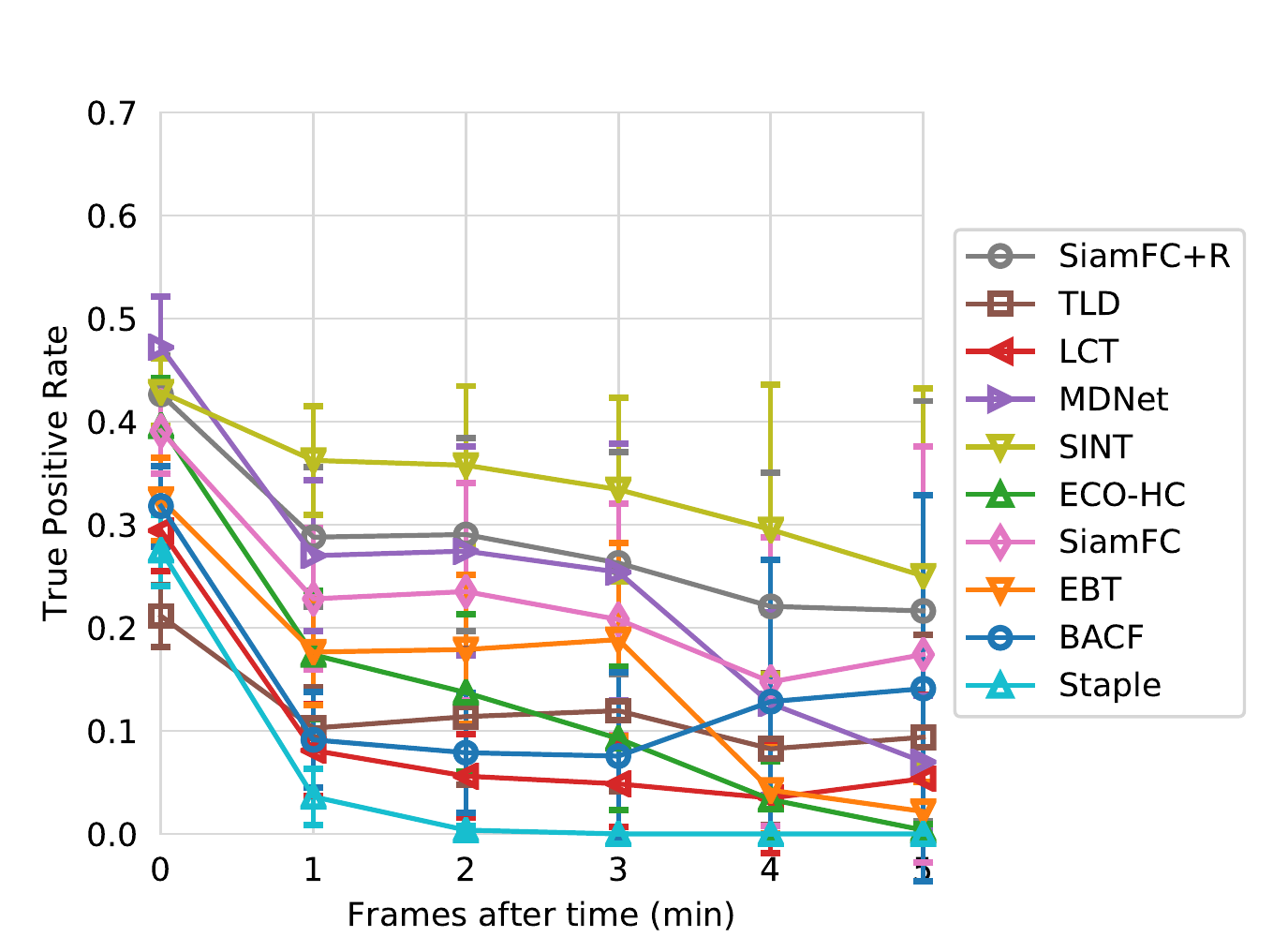}}
\caption{
Degradation of tracker performance over time.
SINT seems more robust to this effect than most other methods.
The variance becomes large when considering only frames beyond four minutes because there are less annotations in this region.
}
\label{fig:tpr-time}
\end{figure}

We analyze the performance of all methods in different time ranges.
Figure~\ref{fig:tpr-time} (left) plots the TPR for frames $t \in (0, x]$ whereas Figure~\ref{fig:tpr-time} (right) plots the TPR for frames $t \in (x, \infty)$.
With the possible exception of SINT, these plots show that the performance of all methods decays rapidly after the first minute.
This seems to be most severe for methods based on online-learned linear templates and hand-crafted features (LCT, Staple, BACF and ECO-HC, to a varying degree).
Although SiamFC is similar in design to SINT, its performance decays more rapidly.
This may be due to architectural differences, or because SINT is initialized with parameters pre-trained for image classification, or because SINT is more restrictive in its scale-space search.

\begin{figure}[t]
\centering
\adjustbox{max width=\linewidth}{
\subfloat{
\includegraphics[height=48mm]{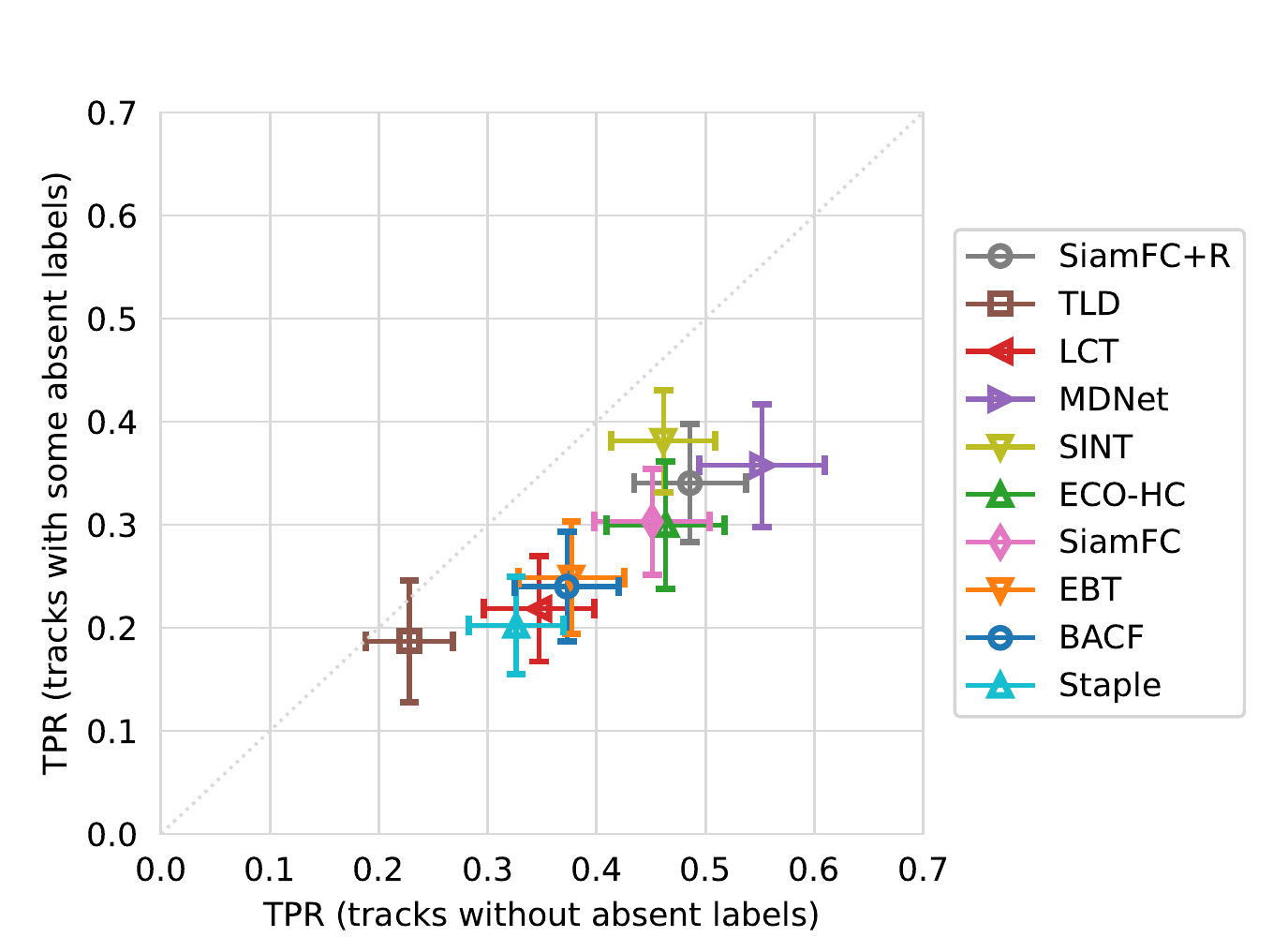}}
\subfloat{
\includegraphics[height=48mm]{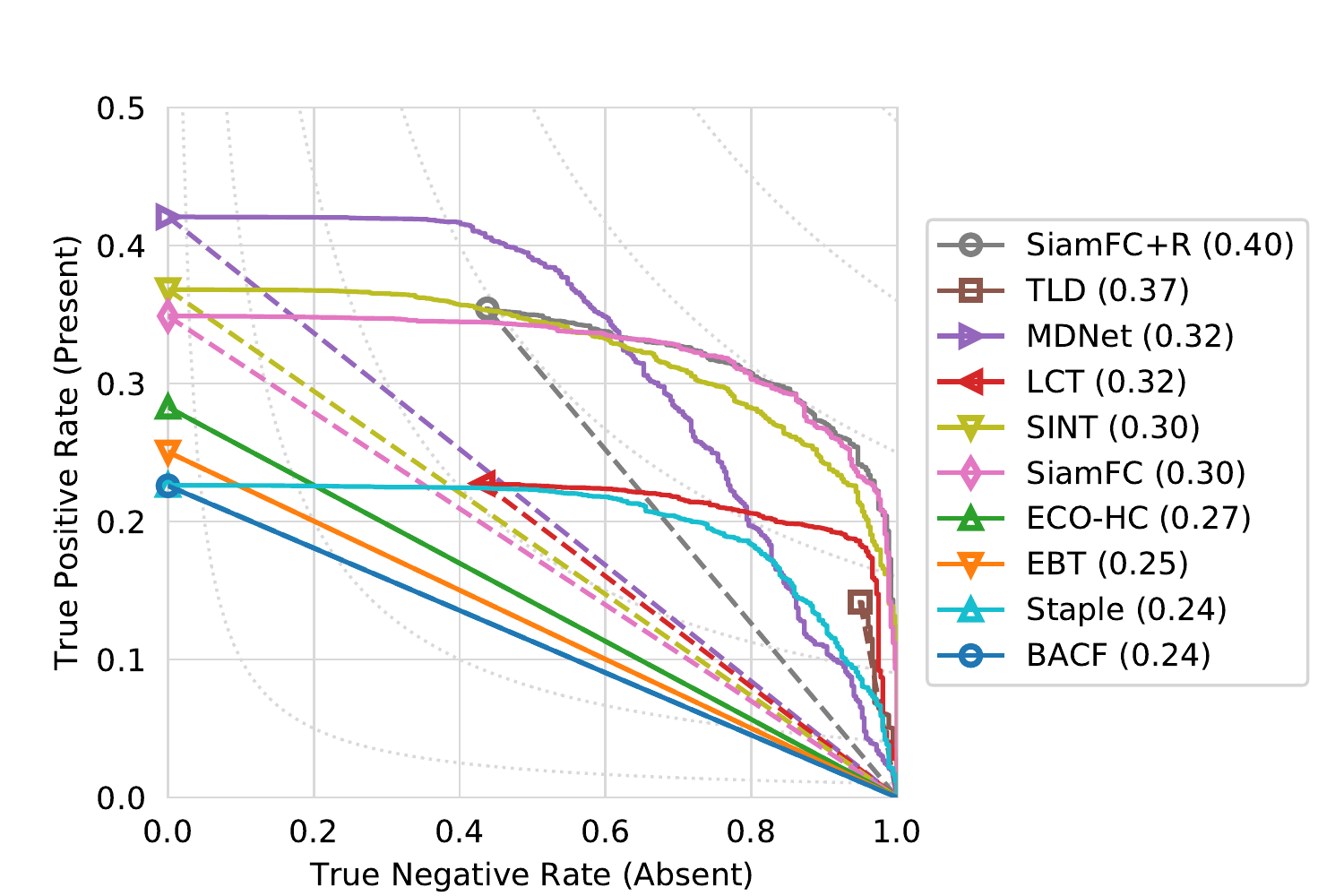}}
}
\caption{\emph{(left)} Impact of disappearances. All baselines are negatively impacted in the presence of target absences, although to a different extent.
\emph{(right)} Effect of post-hoc score thresholding (on the \textit{dev} set) for trackers that output a score.
}
\label{fig:disappearance-posthoc}
\end{figure}

\paragraph{Influence of object disappearance.}
\label{sec:disappearance}
We compare the performance of the different methods on videos that contain at least one \absent{} annotation to those in which every annotation is \present{}.
This is a heuristic for whether the object disappears in the duration of the sequence.
Figure~\ref{fig:disappearance-posthoc} (left) visualizes the relationship between the TPR for these two subsets of videos.
Intuitively, the closer a method is to the diagonal $y = x$, the less its performance is affected by disappearance.

We observe that all baselines have better performance in the set of videos in which the target object never disappears.
This is not surprising, as most methods assume that the target object is always present.
Nonetheless, TLD and SINT seem to be slightly less affected by disappearance than other methods, as they are relatively close to the diagonal.

\paragraph{Post-hoc score thresholding.}
\label{sec:post-hoc}

Although we have stated that we do not wish to evaluate methods at multiple operating points by varying a score threshold, it is natural for a tracker to possess such an internal score, and it may be informative to inspect the result of applying this ``post-hoc'' threshold.
Figure~\ref{fig:disappearance-posthoc} (right) illustrates the different results obtained by sweeping the range of score thresholds.
Note that this plot can only be constructed for the \textit{dev} set, because the evaluation server for the \textit{test} set returns a statistical summary of the results, not the validation of each individual frame.

The large gaps between the lower bound curves (dashed line) and the post-hoc curves (continuous line) show that there is a lot to be gained by simply thresholding the prediction score.
Intuition might suggest that post-hoc thresholding is itself a lower bound on the performance that could be obtained by adjusting the model's internal threshold: if modifying the threshold improves the predictions, then surely it would be even better for the tracker to have made this decision internally?
However, this is not necessarily the case, since changing the internal decision in one frame may have an unpredictable effect in the frames that follow.
Indeed, the re-detection module of SiamFC+R hardly improves over the post-hoc threshold curve of SiamFC.

In the high-TNR region, the approaches based on offline-trained Siamese networks seem more promising than the online-trained MDNet and Staple.

\section{Continuous Attributes}\label{sec:attributes}
\subsection{Definition}
While measuring performance on a large set of videos is an important indicator of a tracker's overall quality, such an aggregate metric hides many subtleties that differentiate trackers. For a more in-depth analysis, modern datasets usually include binary attribute annotations~\cite{smeulders2014visual,wu2015object,kristan2016novel,liang2015encoding,li2016nus,mueller2016benchmark}.
By measuring performance on a subset of videos with a particular attribute, such as ``scale change'' or ``fast motion'', one can characterize the strengths and weaknesses of a tracker.

Unfortunately, the manual annotation of binary attributes is highly subjective: how fast does the target have to move in order to be labelled ``fast motion'', or what is the threshold for ``scale change''?
Instead, we decided to measure quantities that are correlated to some informative attributes, but which can be calculated directly from bounding box annotations and meta-data.
We refer to these quantities as \emph{continuous attributes}.
Each frame $i$ where the target is present is annotated with a time instant $t_{i}$, 2D position vector $p_{i}$, and bounding box dimensions $(w_{i},h_{i})$, expressed as a fraction of the image size.
The continuous attributes are then defined as follows:

\noindent{\bf Size.}
Trackers have different strategies to search across scale,
so they can be sensitive to different object sizes.
The target size at each frame is defined $s_{i}=s(w_{i},h_{i})=\sqrt{w_{i}h_{i}}$.
This metric was chosen because it is invariant to aspect ratio changes (\ie $s(rw_{i}, \allowbreak h_{i}/r)=s(w_{i},h_{i})$). It also changes linearly when the object is re-scaled by an isotropic factor (\ie $s(\sigma w_{i},\sigma h_{i})=\sigma s(w_{i},h_{i})$).
\\{\bf Relative speed.}
Fast-moving targets can lose trackers that depend heavily on temporal smoothness. We compute the target speed relative to its size, $\Delta_{i}$, with:
\[
\Delta_{i}=\frac{1}{\sqrt{s_{i}s_{i-1}}}\frac{\left\Vert p_{i}-p_{i-1}\right\Vert _{2}}{t_{i}-t_{i-1}} \enspace .
\]
The second factor is the instantaneous speed of the target, while the first factor normalizes it w.r.t.\ the object size.
The normalization is needed since the object size is inversely correlated to the distance from the camera, and perspective effects result in closer (larger) objects moving more than objects further away.
\\{\bf Scale~change.}
Some targets may remain mostly at the same scale across a video, while others will vary wildly due to perspective changes.
We measure the range of scale variation in a video as $S=\max_{i}s_{i} / \min_{i}s_{i}$.
\\{\bf Object absence.}
In addition to the analysis of Section~\ref{sec:disappearance}, here we measure performance as a function of the fraction of frames in which the target is \absent{}.
\\{\bf Distractors.} Appearance-based methods can be distracted by objects that are similar to the target, \eg objects of the same class.
To explore this aspect, we leverage the multiple annotations per video and define the number of distractors as the number of other objects with the same class as the target.
\\{\bf Length.} In long videos the effects of small errors are compounded over time, causing trackers to drift.
We define video length as the elapsed time in seconds between the first and last annotations of the target.

While these attributes could be thresholded to yield binary attributes that are comparable to the previous benchmarks, we found that \emph{binning} them can yield a more informative plot, especially if performance is presented together with the size of each bin, in order to indicate its reliability.

\begin{figure}[t]
\centering
\includegraphics[width=0.95\textwidth]{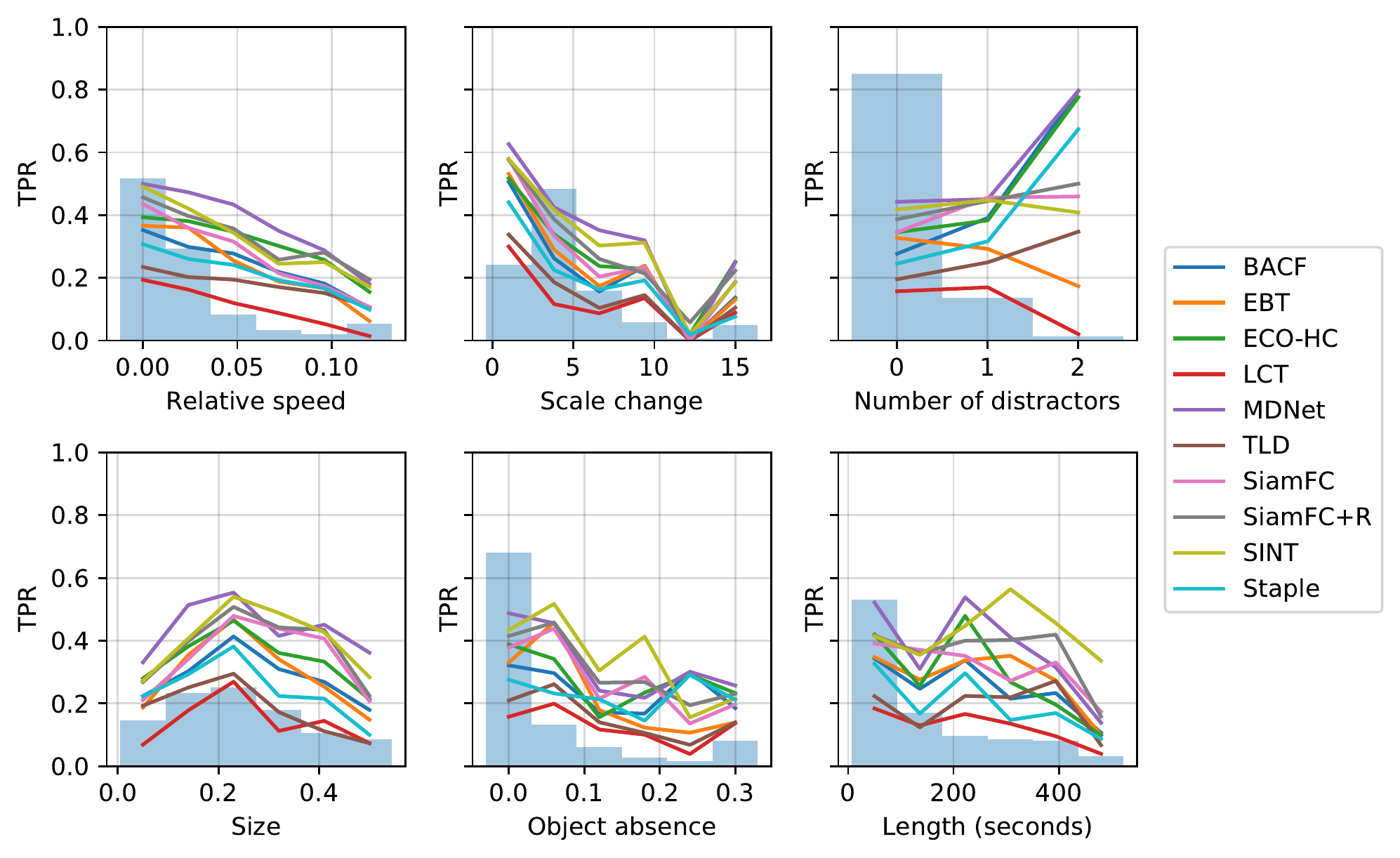}
\caption{True-Positive Rate (at IOU $\ge 0.5$) of each tracker as a function of different continuous attributes. The continuous attributes are computed per frame or per video, and are then distributed into discrete bins. TPR is computed separately for the frames/videos in each bin. The shaded boxes show the fraction of frames/videos that belong to each bin. Only relative speed and size are computed per-frame, the remaining are per-video.}\label{fig:attributes}
\end{figure}

\subsection{Influence of Continuous Attributes}
\label{sec:attributes-experiments}

We partition each attribute into 6 bins, except for ``distractors'' which takes only 3 discrete values.
Fig.~\ref{fig:attributes} shows a histogram (shaded boxes) with the fraction of frames/videos that fall into each bin of each attribute, together with a plot indicating the performance (TPR) of each tracker over the subset corresponding to each bin.
Notice how, for the points in the plots corresponding to bins with fewer videos, the variance is quite high and thus their results may be difficult to interpret.
However, we can still draw several conclusions:
\\{\bf Relative speed.}
Unsurprisingly, all trackers performing local search show degraded performance as the target moves more rapidly.
Among all the methods able to consider the entire frame (TLD, LCT, EBT and SiamFC+R) the least affected by high speeds is TLD.
\\{\bf Scale change.} Videos where the target maintains the same size seem to be the optimal operating point for all trackers. There is a significant dip in performance around 6$\times$ variation in scale. Since this bin contains a significant fraction of the videos, there is a large opportunity for improvement by focusing on this case.
\\{\bf Number of distractors.} Methods do not seem to be confounded by distractors of the same class as much as one would expect.
For videos with one distractor, most trackers' performances are maintained.
This means that they are not simply detecting broad object categories, which was a plausible concern over the use of pre-trained deep networks.
With two distractors, only EBT and LCT seem to perform significantly worse, possibly locking on distractor objects during their full-image search strategy.
\\{\bf Size.}
Most trackers seem to be well-adapted to the range of object sizes in the dataset, with a performance peak reached at 0.2 by all the methods taken into account. Unlike others, MDNet and LCT seem to maintain their performances at the largest object sizes.
\\{\bf Object absence.}
As already noted in Section~\ref{sec:disappearance}, disappearance of the target object affects all methods, which show a meaningful drop in performance when the number of frames where the object is \absent{} increases from 0\% to 10\%.
SiamFC+R, MDNet and ECO-HC seem to be less affected by larger absences.
\\{\bf Length.} As noted in Section~\ref{sec:video_length}, probably due to the short-term nature of the benchmarks that they were calibrated for, most trackers are severely affected after only a few minutes of tracking.
For example, both MDNet and ECO-HC present a large drop in performance in videos longer than three minutes. 
SINT, followed by MDNet, are the most obvious exceptions to this trend.

\section{Conclusion}\label{sec:conclusions}
We have introduced the OxUvA long-term tracking dataset, with which it is possible to assess methods on sequences that are minutes in length and often contain disappearance of the target object.
Our benchmark is the largest ever proposed in the single-object tracking community and contains more than $25\times$ the number of frames of OTB-100.
In order to afford such a vast dataset, we opt for a relatively sparse labelling of the target objects at 1Hz.
To justify this decision, we empirically show that, for the sake of reliability, a high density of labels is not important while a large number of videos is paramount.

Adapting the metrics of True Positive and True Negative Rate from classification, we design an evaluation that measures the ability of a tracker to correctly understand whether the target object is present in the frame and where it is located.
We then evaluate the performance of several popular tracking methods on the 166 sequences that comprise our testing set, also considering the effect of several factors such as the object's speed and size, the sequence length, the number of distractors and the amount of occlusion.
We believe that our contribution will spur the design of algorithms ready to be used in the many practical applications that require trackers able to deal with long sequences and capable of determining whether the object is present or not.

\bibliographystyle{splncs04}
\bibliography{main}

\clearpage

\appendix

\section{Dev and Test Subsets}

\begin{figure}
\centering
\begin{tabular}{l l}
\multicolumn{1}{c}{\textbf{Dev}} & \multicolumn{1}{c}{\textbf{Test}} \\
\includegraphics[height=45mm]{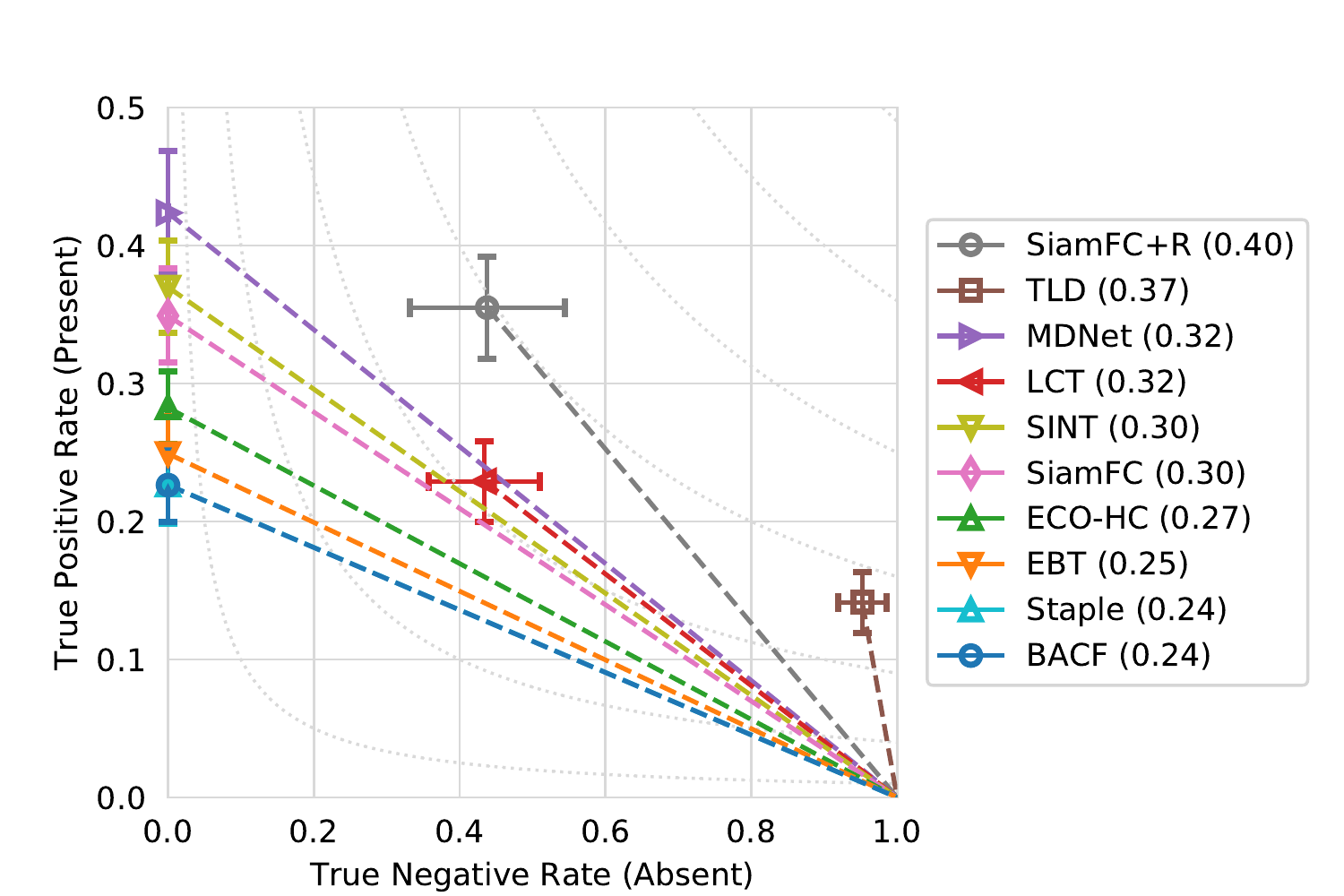} &
\includegraphics[height=45mm]{figures/analysis/test/open/tpr_tnr_iou_0d5_bootstrap} \\
\includegraphics[height=45mm]{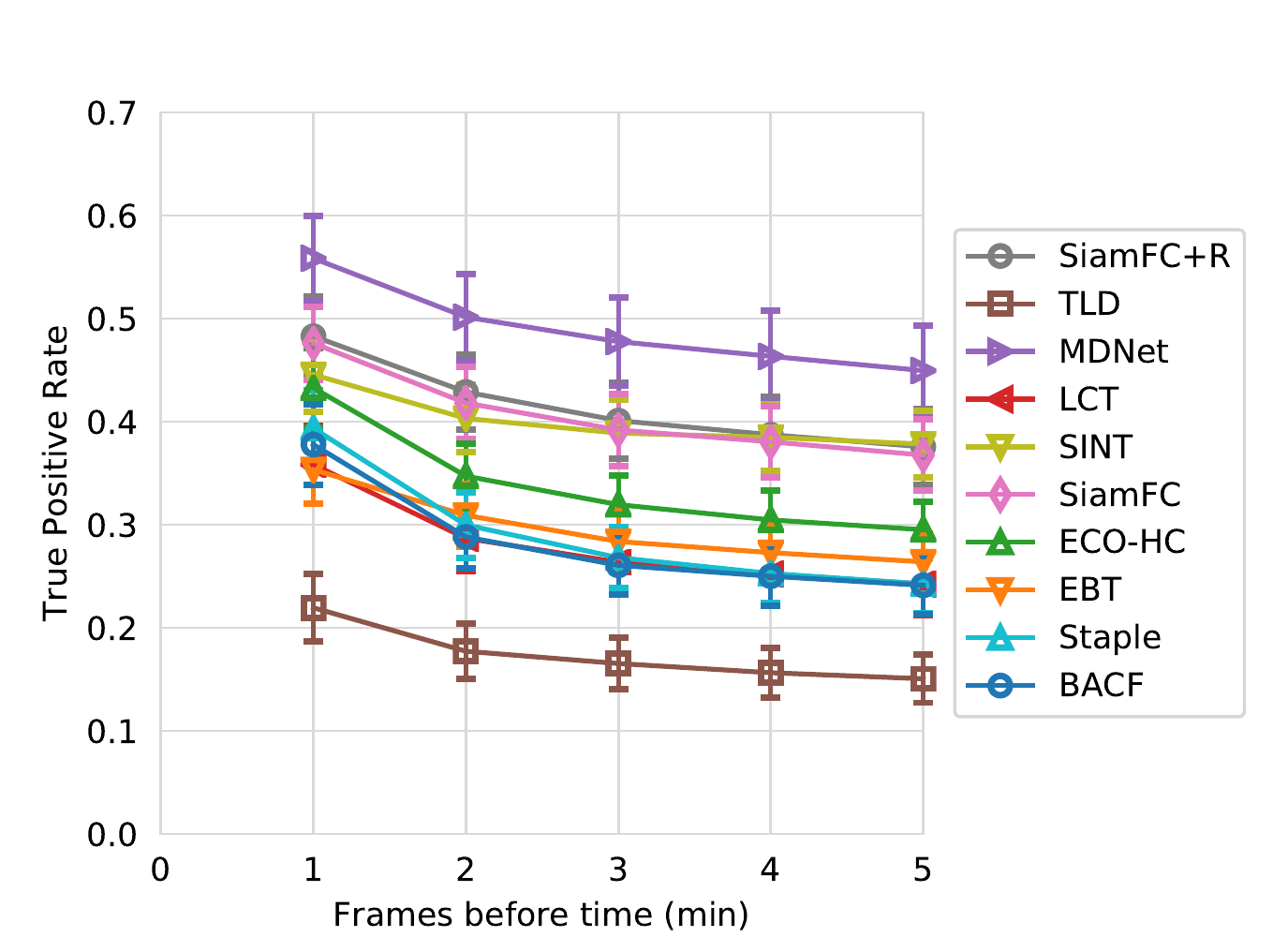} &
\includegraphics[height=45mm]{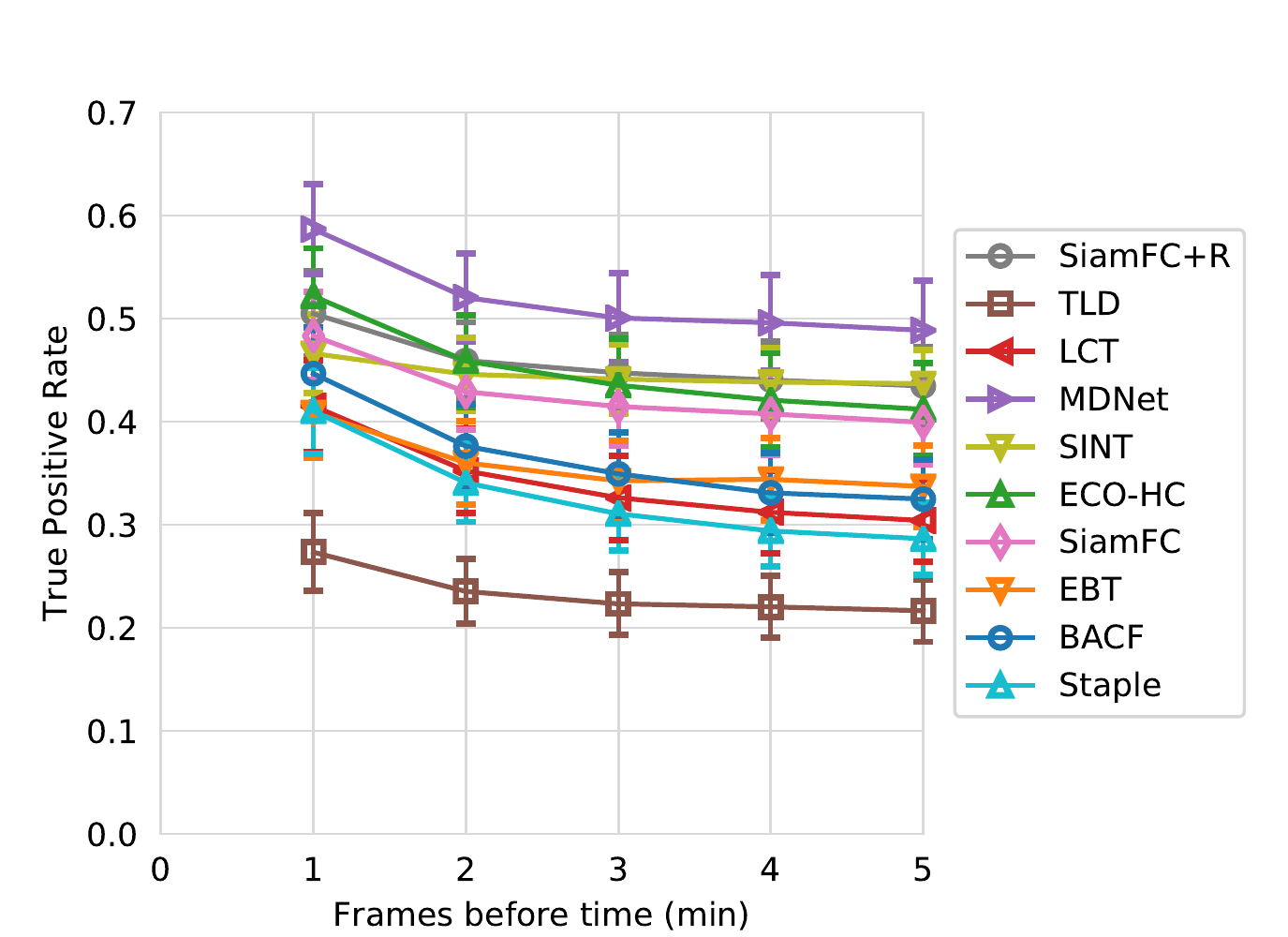} \\
\includegraphics[height=45mm]{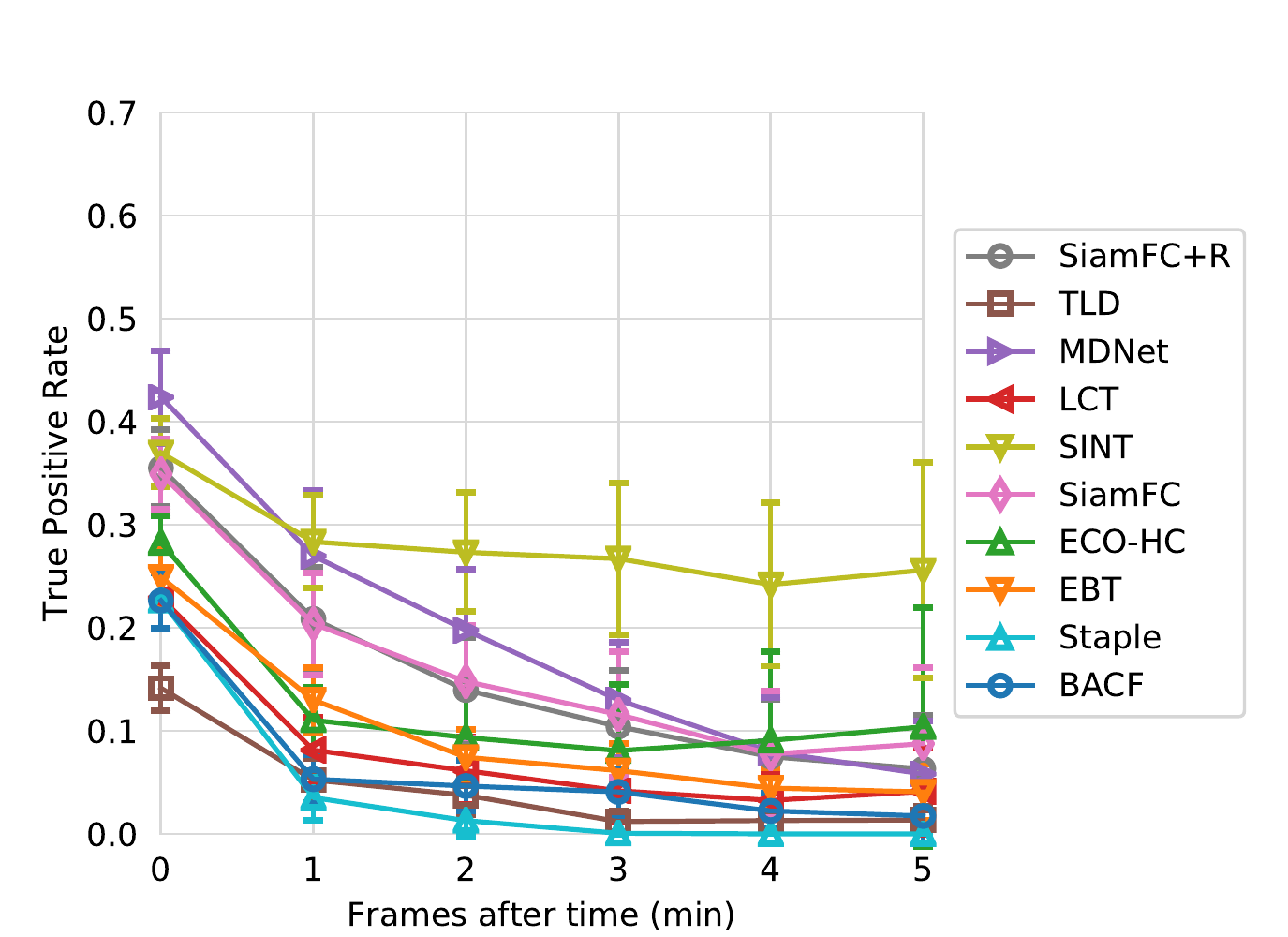} &
\includegraphics[height=45mm]{figures/analysis/test/open/tpr_time_iou_0d5_interval_after_bootstrap} \\
\end{tabular}
\caption{The main plots for the dev set compared to those for the test set.}
\label{fig:subsets}
\end{figure}

All results presented in the main paper use the test set (except for the post-hoc thresholding experiment).
Figure~\ref{fig:subsets} presents, side by side, the main plots for both the dev and test sets.
It can be seen that the values for TPR and TNR are significantly different.
However, this is expected since the two subsets contain disjoint classes, and therefore the distributions are different.
Nevertheless, the relative trends between trackers are preserved.

\section{Constrained and Open Challenges}

\begin{figure}
\centering
\begin{tabular}{l l}
\multicolumn{1}{c}{\textbf{Constrained}} & \multicolumn{1}{c}{\textbf{Open}} \\
\includegraphics[height=45mm]{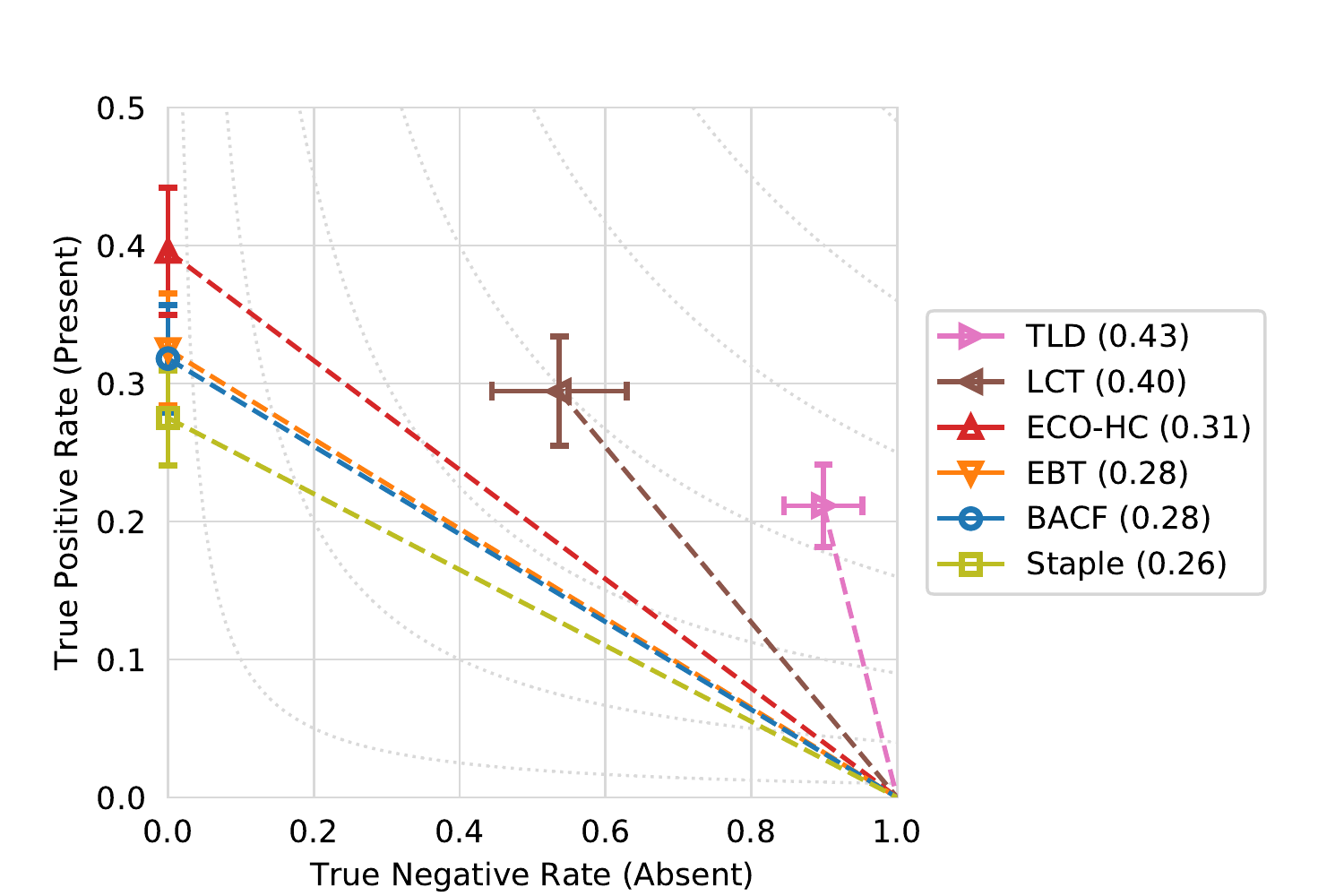} &
\includegraphics[height=45mm]{figures/analysis/test/open/tpr_tnr_iou_0d5_bootstrap} \\
\includegraphics[height=45mm]{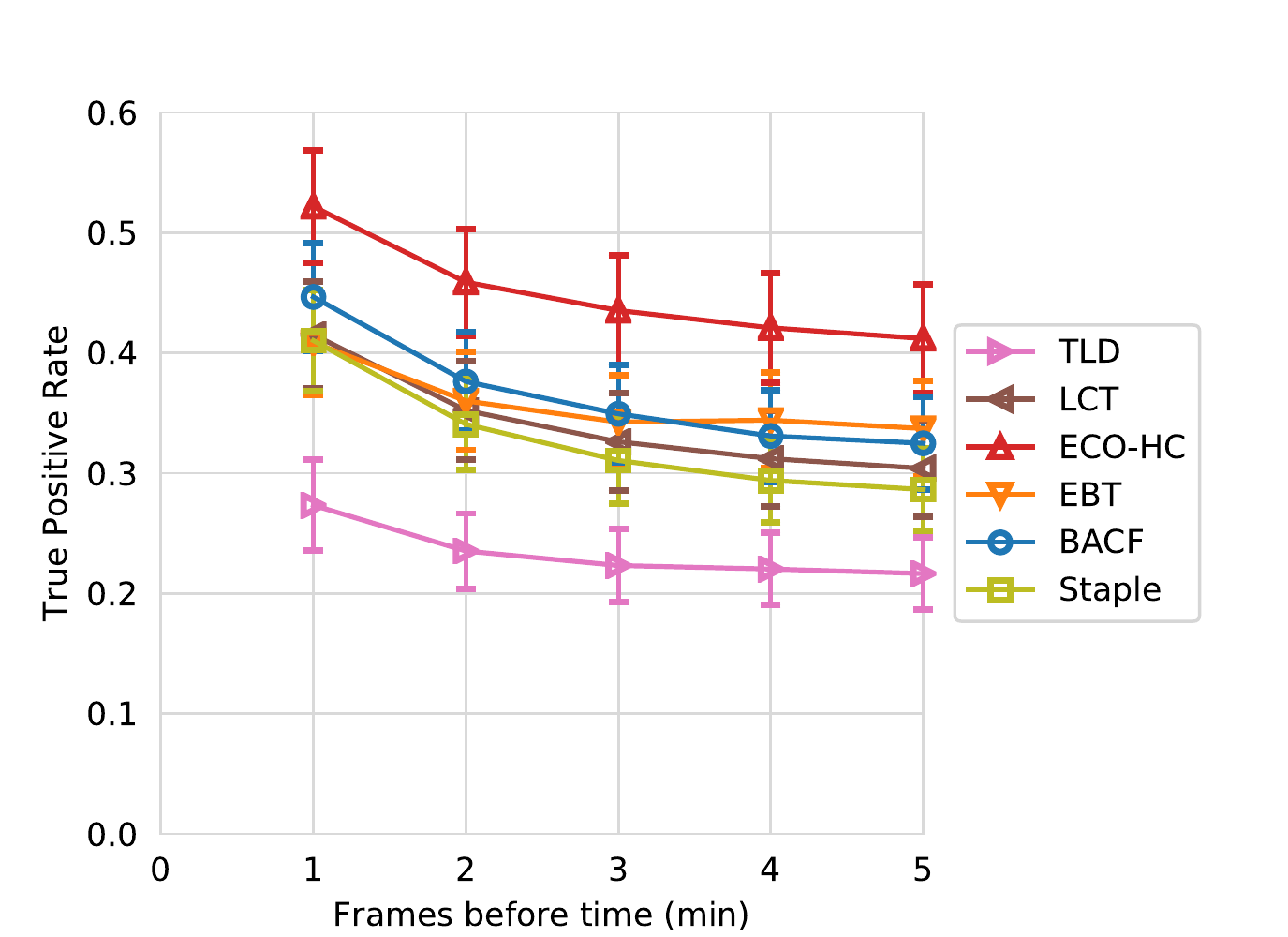} &
\includegraphics[height=45mm]{figures/analysis/test/open/tpr_time_iou_0d5_interval_before_bootstrap} \\
\includegraphics[height=45mm]{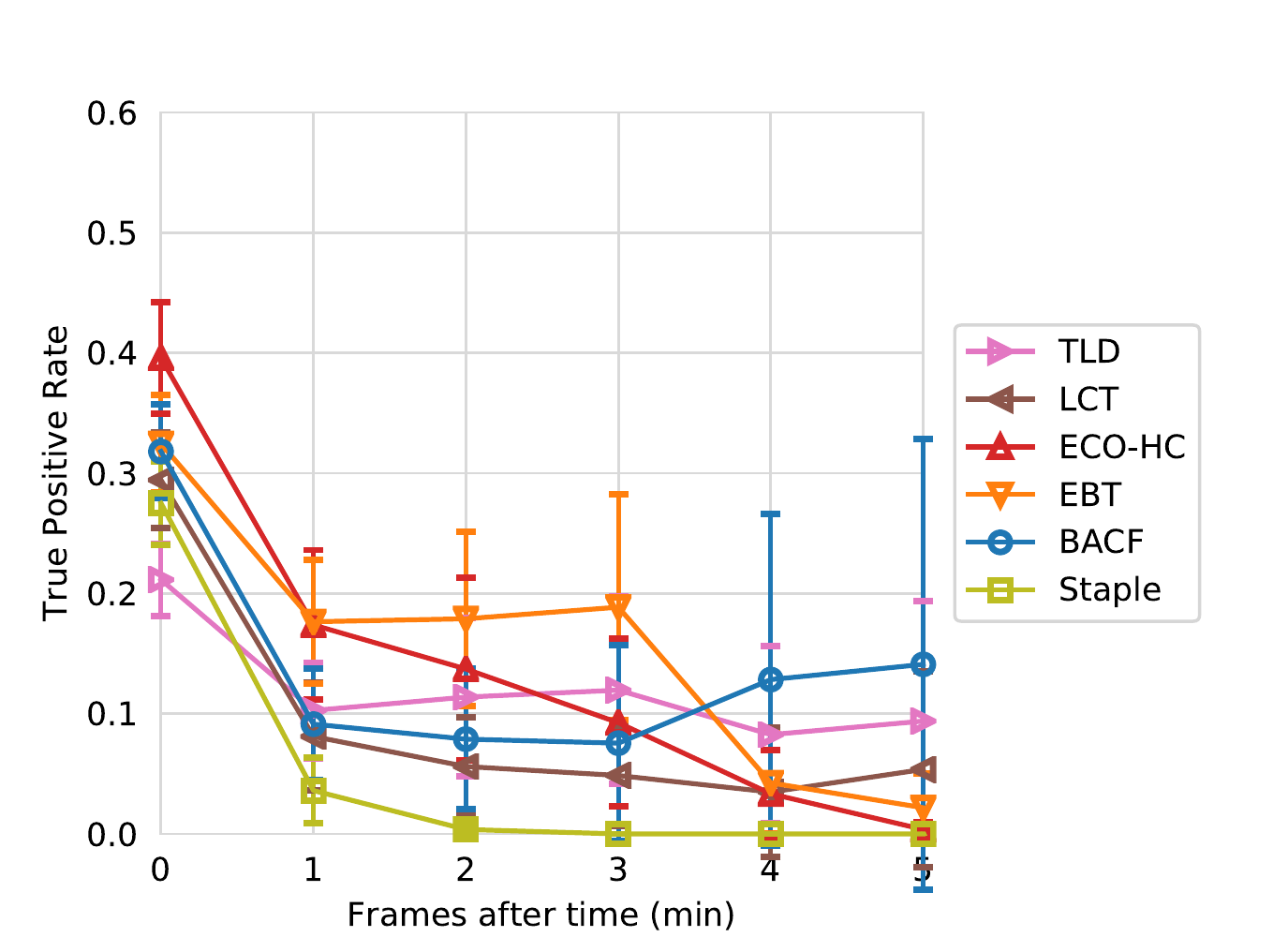} &
\includegraphics[height=45mm]{figures/analysis/test/open/tpr_time_iou_0d5_interval_after_bootstrap} \\
\end{tabular}
\caption{Only a subset of trackers are eligible for the constrained challenge.}
\label{fig:challenges}
\end{figure}

The main paper presents results for all trackers.
However, a subset of trackers are eligible for the constrained challenge.
Figure~\ref{fig:challenges} shows the main plots for the constrained challenge beside their equivalent from the main paper.
Excluding the trackers based on deep conv-nets, which either use ImageNet VID or pre-trained weights, TLD obtains the best score.
However, the other methods are at a serious disadvantage because they never predict \absent{}.
The best methods that assume the object is always present are ECO-HC and EBT.

\section{Object Classes}

Similarly to standard tracking benchmarks, we do not make the class labels available to the tracking algorithms during development or testing.
However, we believe it is interesting to know the variety of target object classes in a dataset.
For this reason, besides reporting the class labels for OxUvA, we attempt to present the same information for NUS-PRO, OTB-100, TC, UAV123 and VOT-2017, clustering object categories into (coarsely defined) classes and counting the number of instances per class.
The statistics are reported in Table~\ref{tab:classes}.

While other benchmarks place a strong emphasis on few classes, our sequences are more equally distributed.
For example, the most frequent class in our dataset (\textit{bear}) only occurs in 14\% of the videos.
Instead, the class \textit{person}, which is the most frequent in all the other benchmarks, appears in at least 35\% and up to 53\% of the videos.

\begin{sidewaystable}
    \centering
    \scriptsize
    \begin{tabular}{ccc ccc ccc ccc ccc cc}
        \toprule
        \multicolumn{2}{c}{\scriptsize{NUS-PRO~\cite{li2016nus}}} &  & \multicolumn{2}{c}{\scriptsize{OTB-100~\cite{wu2015object}}} & & \multicolumn{2}{c}{\scriptsize{TC~\cite{liang2015encoding}}} & & \multicolumn{2}{c}{\scriptsize{UAV123~\cite{mueller2016benchmark}}} & & \multicolumn{2}{c}{\scriptsize{VOT-17~\cite{Kristan_2017_ICCV_Workshops}}} & & \multicolumn{2}{c}{\scriptsize{Ours}}\\
        class & \# entries && class & \# entries && class & \# entries && class & \# entries && class & \# entries && class & \# entries \\
        \midrule
        \emph{person} & 193 && \emph{person} & 34 && \emph{person} & 45 && \emph{person} & 48 && \emph{person} & 19 && \emph{bear} & 50 \\
        \emph{head} & 60 && \emph{head} & 26 && \emph{head} & 16 && \emph{car} & 30 && \emph{head} & 5 && \emph{person} & 35 \\
        \emph{car} & 31 && \emph{car} & 12 && \emph{sphere} & 8 && \emph{drone} & 10 && \emph{fish} & 4 && \emph{bird} & 32 \\
        \emph{airplane} & 20 && \emph{toy} & 8 && \emph{2D print} & 5 && \emph{wakeboard} & 10 && \emph{motorcycle} & 4 && \emph{dog} & 31 \\
        \emph{boat} & 20 && \emph{2D print} & 4 && \emph{bicycle} & 5 && \emph{boat} & 9 && \emph{car} & 3 && \emph{boat} & 27 \\
        \emph{helicopter} & 20 && \emph{cuboid} & 3 && \emph{car} & 5 && \emph{building} & 5 && \emph{drone} & 3 && \emph{horse} & 27 \\
        \emph{motorcycle} & 20 && \emph{bird} & 2 && \emph{ball} & 4 && \emph{truck} & 5 && \emph{ant} & 2 && \emph{elephant} & 25 \\
        \emph{drone} & 1 && \emph{motorcycle} & 1 && \emph{toy} & 4 && \emph{bicycle} & 3 && \emph{ball} & 2 && \emph{airplane} & 22 \\
        - & - && \emph{deer} & 1 && \emph{hand} & 3 && \emph{bird} & 3 && \emph{bird} & 2 && \emph{skateboard} & 21 \\
        - & - && \emph{bottle} & 1 && \emph{kite} & 3 && - & - && \emph{toy} & 2 && \emph{knife} & 20 \\
        - & - && \emph{panda} & 1 && \emph{logo} & 3 && - & - && \emph{bag} & 1 && \emph{bus} & 14 \\
        - & - && \emph{board} & 1 && \emph{cuboid} & 3 && - & - && \emph{book} & 1 && \emph{truck} & 11 \\
        - & - && \emph{can} & 1 && \emph{boat} & 2 && - & - && \emph{butterfly} & 1 && \emph{bicycle} & 10 \\
        - & - && \emph{dog} & 1 && \emph{cup} & 2 && - & - && \emph{cable} & 1 && \emph{car} & 10 \\
        - & - && \emph{transformer} & 1 && \emph{fish} & 2 && - & - && \emph{crab} & 1 && \emph{umbrella} & 8 \\
        - & - && \emph{bicycle} & 1 && \emph{guitar} & 2 && - & - && \emph{cat} & 1 && \emph{cow} & 6 \\
        - & - && - & - && \emph{bird} & 2 && - & - && \emph{flamingo} & 1 && \emph{potted plant} & 6 \\
        - & - && - & - && \emph{microphone} & 2 && - & - && \emph{frisbee} & 1 && \emph{motorcycle} & 4 \\
        - & - && - & - && \emph{torso} & 2 && - & - && \emph{glove} & 1 && \emph{toilet} & 3 \\
        - & - && - & - && \emph{motorcycle} & 2 && - & - && \emph{hand} & 1 && \emph{giraffe} & 2 \\
        - & - && - & - && \emph{airplane} & 2 && - & - && \emph{helicopter} & 1 && \emph{cat} & 1 \\
        - & - && - & - && \emph{board} & 1 && - & - && \emph{leaf} & 1 && \emph{zebra} & 1 \\
        - & - && - & - && \emph{bottle} & 1 && - & - && \emph{rabbit} & 1 && - & - \\
        - & - && - & - && \emph{can} & 1 && - & - && \emph{sheep} & 1 && - & - \\
        - & - && - & - && \emph{deer} & 1 && - & - && - & - && - & - \\
        - & - && - & - && \emph{ring} & 1 && - & - && - & - && - & - \\
        - & - && - & - && \emph{torus} & 1 && - & - && - & - && - & - \\
        \bottomrule
    \end{tabular}
    \caption{Occurrence of target object classes in popular object tracking benchmarks.}
    \label{tab:classes}
\end{sidewaystable}

\end{document}